\documentclass{article} 
\usepackage{iclr2017_conference,times}
\usepackage{hyperref}
\usepackage{url}
\usepackage{amsmath}
\usepackage{graphicx}
\usepackage{algorithm}
\usepackage{multirow}
\usepackage{comment}
\usepackage[noend]{algpseudocode}
\usepackage{subfig}
\usepackage[utf8]{inputenc}

\title{Prediction Under Uncertainty with Error-Encoding Networks}

\author{Mikael Henaff, Junbo Zhao and Yann LeCun \\
Facebook AI Research \\
Courant Institute, New York University}

\iclrfinalcopy 
\begin{document}

\maketitle

\begin{abstract}
In this work we introduce a new framework for performing temporal predictions in the presence of uncertainty. It is based on a simple idea of disentangling components of the future state which are predictable from those which are inherently unpredictable, and encoding the unpredictable components into a low-dimensional latent variable which is fed into a forward model. Our method uses a supervised training objective which is fast and easy to train. We evaluate it in the context of video prediction on multiple datasets and show that it is able to consistently generate diverse predictions without the need for alternating minimization over a latent space or adversarial training.
\end{abstract}

\section{Introduction}
Learning forward models in time series is a central task in artificial intelligence, with applications in unsupervised learning, planning and compression. 
A major challenge in this task is how to handle the multi-modal nature of many time series. 
When there are multiple valid ways in which a time series can evolve, training a model using classical $\ell_1$ or $\ell_2$ losses produces predictions which are the average or median of the different outcomes across each dimension, which is itself often not a valid prediction. 

In recent years, Generative Adversarial Networks \citep{GANs} have been introduced, a general framework where the prediction problem is formulated as a minimax game between the predictor function and a trainable discriminator network representing the loss. By using a trainable loss function, it is in theory possible to handle multiple output modes since a generator which covers each of the output modes will fool the discriminator leading to convergence. However, a generator which covers a single mode can also fool the discriminator and converge, and this behavior of mode collapse has been widely observed in practice. Some workarounds have been introduced to resolve or partially reduce mode-collapsing, such as minibatch discrimination, adding parameter noise \citep{Salimans2016}, backpropagating through the unrolled discriminator \citep{Metz16} and using multiple GANs to cover different modes \citep{AdaGAN}. However, many of these techniques can bring additional challenges such as added complexity of implementation and increased computational cost. The mode collapsing problem becomes even more pronounced in the conditional generation setting when the output is highly dependent on the context, such as video prediction \citep{Mathieu15, Isola2016}. 


In this work, we introduce a novel architecture that allows for robust multimodal conditional predictions in time series data.
It is based on a simple intuition of separating the future state into a deterministic component, which can be predicted from the current state, and a stochastic (or difficult to predict) component which accounts for the uncertainty regarding the future mode.
By training a model deterministically, we can obtain this factorization in the form of the model's prediction together with the prediction error with respect to the true state. This error can be encoded as a low-dimensional latent variable which is fed back into the model to accurately correct the determinisic prediction by incorporating this additional information.
We call this model the Error Encoding Network (EEN).
In a nutshell, this framework contains three function mappings at each timestep: (i) a mapping from the current state to the future state, which separates the future state into deterministic and non-deterministic components; (ii) a mapping from the non-deterministic component of the future state to a low-dimensional latent vector; (iii) a mapping from the current state to the future state conditioned on the latent vector, which encodes the mode information of the future state.
While the training procedure involves all these mappings, the inference phase involves only (iii). 

The model is trained end-to-end using a supervised learning objective and latent variables are computed using a learned parametric function, leading to easy and fast training.
We apply this method to video datasets from games, robotic manipulation and simulated driving, and show that the method is able to consistently produce multimodal predictions of future video frames for all of them.
Although we focus on video in this work, the method itself is general and can in principle be applied to any continuous-valued time series.

\section{Model}

Many natural processes carry some degree of uncertainty. This uncertainty may be due to an inherently stochastic process, a deterministic process which is partially observed, or it may be due to the complexity of the process being greater than the capacity of the forward model. 
One natural way of dealing with uncertainty is through latent variables, which can be made to account for aspects of the target that are not explainable from the observed input. 

Assume we have a set of continuous vector-valued input-target pairs $(x_i, y_i)$, where the targets depend on both the inputs and some inherently unpredictable factors. 
For example, the inputs could be a set of consecutive video frames and the target could be the following frame.
Classical latent variable models such as $k$-means or mixtures of Gaussians are trained by alternately minimizing the loss with respect to the latent variables and model parameters; in the probabilistic case this is the Expectation-Maximization algorithm \citep{EM}. 
In the case of a neural network model $f_{\theta}(x_i, z)$, continuous latent variables can be optimized using gradient descent and the model can be trained with the following procedure:

\begin{algorithm}
\caption{Train latent variable model with alternating minimization}\label{euclid}
\begin{algorithmic}[1]
\Require Learning rates $\alpha, \beta$, number of iterations $K$.
\Repeat
\State Sample $(x_i, y_i)$ from the dataset
\State initialize $z \sim \mathcal{N}(0, 1)$
\State $i \leftarrow 1$
\While{$i \leq K$}
\State $z \leftarrow z - \alpha \nabla_z \mathcal{L}(y_i, f_{\theta}(x_i, z))$
\State $i \leftarrow i + 1$
\EndWhile
\State $\theta \leftarrow \theta - \beta \nabla_\theta \mathcal{L}(y_i, f_{\theta}(x_i, z))$
\Until converged
\end{algorithmic}
\end{algorithm}

Our approach is based on two observations. First, the latent variable $z$ should represent what is not explainable using the input $x_i$. 
Ideally, the model should make use of the input $x_i$ and only use $z$ to account for what is not predictable from it. 
Second, if we are using gradient descent to optimize the latent variables, $z$ will be a continuous function of $x_i$ and $y_i$, although a possibly highly nonlinear one. 

Our model has two settings: a deterministic setting, where it produces a prediction using only $x_i$, and a conditional setting where is produces a prediction using $x_i$ and a latent variable $z$. We can switch to the deterministic setting by fixing $z=0$; optionally, we can also have a separate network or set of weights for each setting. 
We first train the model $f_\theta(x, z)$ in the deterministic setting to minimize the following loss over the training set:

\begin{equation}
\mathcal{L}_d(\theta) = \sum_i \| y_i - f_{\theta}(x_i, 0) \|
\end{equation}

Here the norm can denote $\ell_1$, $\ell_2$ or any other loss which is a function of the difference between the target and the prediction. Given sufficient data and capacity, $f$ will learn to extract all the information possible about each $y_i$ from the corresponding $x_i$, and what is inherently unpredictable will be contained within the residual error, $y_i - f_{\theta}(x_i, 0)$. 

Once $f$ is fully trained in the deterministic setting, we save a copy of the parameters $\theta_{-}$ and then continue training by minimizing the following loss over the training data:

\begin{equation}
\mathcal{L}_c(\theta, \phi) = \sum_i \| y_i - f_{\theta}(x_i, \phi(y_i-f_{\theta_{-}}(x_i, 0)) \|
\end{equation}

Here, $\phi$ is a learned parametric function which maps the residual error of the model in its deterministic setting to a low-dimensional latent variable $z$ which encodes the identity of the mode to which the future state belongs. This is then used as input to $f$ in its conditional setting to more accurately predict $y_i$, conditioned on knowledge of the proper mode. 
For each sample, we perform two passes through $f$: a first pass on the deterministic setting with $z=0$ and using the parameters $\theta_{-}$ which minimize (1) to compute the residual error which will be input to $\phi$, and a second pass on the conditional setting using the output of $\phi$ as $z$ and the current set of parameters $\theta$. 

The fact that $z$ is a function of the residual prediction error $y_i - f_{\theta_{-}}(x_i, 0)$ reflects the intuition that it should only account for what is not explainable by the input, while still being a continuous function of $x_i$ and $y_i$. Note that using a copy of previous weights $\theta_{-}$ helps prevent information that could be predicted from $x_i$ from being stored in $z$, which could happen if we used the current weights $\theta$ which may become different from $\theta_{-}$ over time. As an alternative, we could use a single set of weights and keep minimizing $\mathcal{L}_d$ jointly with $\mathcal{L}_c$ to prevent this from happening. We tried both methods and found that using a previous version of the weights worked better in some cases.

\begin{figure}
  \centering
 \vspace{-3.0cm}
    \includegraphics[width=\textwidth]{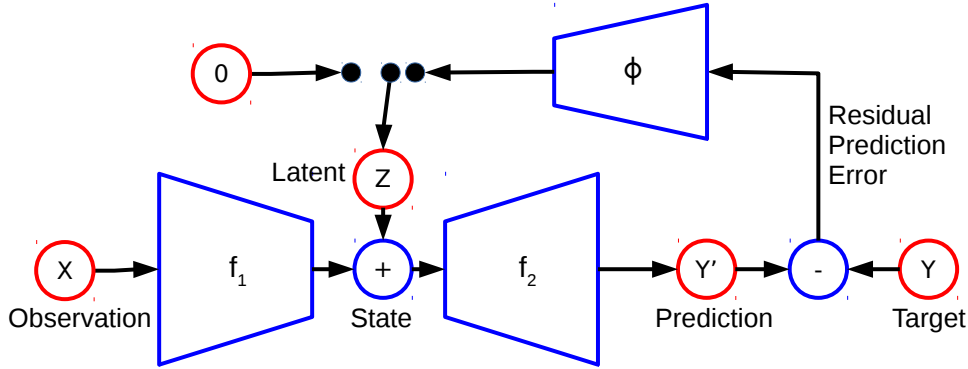}
 \vspace{-3.0cm}
  \caption{Model Architecture. The switch changes between the deterministic setting where $z=0$ and the conditional setting where $z$ is a latent variable representing the inherently unpredictable aspects of the target. The switch can also change the parameters used in the encoder and decoder.}
  \label{architecture}
\end{figure}

The model architecture is shown in Figure \ref{architecture}. In our experiments, we used the architecture $f_\theta(x, z) = f_2(f_1(x) + Wz)$, where $f_1$ and $f_2$ are the encoder and decoder of the state respectively. 
Note that $z$ is typically of much lower dimension than the residual error $y_i - f_{\theta_{-}}(x_i, 0)$, which prevents the network from learning a trivial solution where $f$ would simply invert $\phi$ and cancel the error from the prediction. This forces the $\phi$ network to map the errors to general representations which can be reused across different samples and correspond to different modes of the conditional distribution. 

To perform inference after the network is trained,  we first extract and save the $z_i = \phi(y_i - f_{\theta_{-}}(x_i, 0))$ from each sample in the training set. 
Given some new input $x'$, we can then generate different predictions by computing $f_{\theta}(x', z')$, for different $z' \in \{z_i\}$. 
In this work, we adopt a simple strategy of sampling uniformly from this set to generate new samples, however more sophisticated methods could be used such as fitting a conditional distribution over $p(z|x)$ and sampling from it. 

\section{Related Work}

In recent years a number of works have explored video prediction. These typically train models to predict future frames with the goal of learning representations which disentangle factors of variation and can be used for unsupervised learning \citep{Srivastava15, Villegas17, DentonB17}, or learn action-conditional forward models which can be used for planning \citep{Oh15, FinnGL16, Poke, VideoPixel}. In the first case, the predictions are deterministic and ignore the possibly multimodal nature of the time series. In the second, it is possible to make different predictions about the future by conditioning on different actions, however this requires that the training data includes additional action labels. Our work makes different predictions about the future by instead conditioning on latent variables which are extracted in an unsupervised manner from the videos themselves.

Several works have used adversarial losses in the context of video prediction. 
The work of \citep{Mathieu15} used a multiscale architecture and a combination of several different losses to predict future frames in natural videos. They found that the addition of the adversarial loss and a gradient difference loss improved the generated image quality, in particular by reducing the blur effects which are common when using $\ell_2$ loss. However, they also note that the generator learns to ignore the noise and produces similar outputs to a deterministic model trained without noise. This observation was also made by \citep{Isola2016} when training conditional networks to perform image-to-image translation. 

Other works have used models for video prediction where latent variables are inferred using alternating minimization. The model in \citep{VondrickPT15} includes a discrete latent variable which was used to choose between several different networks for predicting hidden states of future video frames obtained using a pretrained network. This is more flexible than a purely deterministic model, however the use of a discrete latent variable still limits the possible future modes to a discrete set. The work of \citep{Goroshin15} also made use of latent variables to model uncertainty, which were inferred through alternating minimization. In contrast, our model infers continuous latent variables through a learned parametric function. This is related to algorithms which learn to predict the solution of an iterative optimization procedure \citep{LISTA}. 

Recent work has shown that good generative models can be learned by jointly learning representations in a latent space together with the parameters of a decoder model \citep{GLO}. This leads to easier training than adversarial networks. This generative model is also learned by alternating minimization over the latent variables and parameters of the decoder model, however the latent variables for each sample are saved after each update and optimization resumes when the corresponding sample is drawn again from the training set. This is related to our method, with the difference that rather than saving the latent variables for each sample we compute them through a learned function of the deterministic network's prediction error. 

Our work is related to predictive coding models \citep{rao1999pcv, Spratling2008, DeepPC, PredNet} and chunking architectures \citep{SchmidhuberChunker}, which also pass residual errors or incorrectly predicted inputs between different parts of the network. It differs in that these models pass errors upwards to higher layers in the network at each timestep, whereas our method passes the compressed error signal from the deterministic model backwards in time to serve as input for the model in its conditional setting at the previous timestep. 

\section{Experiments}

We tested our method on five different video datasets from different areas such as games (Atari Breakout, Atari Seaquest and Flappy Bird), robot manipulation \citep{Poke} and simulated driving \citep{Jiakai2016}. 
These have a well-defined multimodal structure, where the environment can change due to the actions of the agent or other stochastic factors and span a diverse range of visual environments.
For each dataset, we trained our model to predict the following 1 or 4 frames conditioned on the previous 4 frames. We also trained a deterministic baseline model and a GAN to compare performance. 
Code to train our models and obtain video generations is available at \url{https://github.com/mbhenaff/EEN}.

The deterministic model and EEN were trained using the $\ell_2$ loss for all datasets except the Robot dataset, where we found that the $\ell_1$ loss gave better-defined predictions. Although more sophisticated losses exist, such as the Gradient Difference loss \citep{Mathieu15}, our goal here was to evaluate whether our model could capture multimodal structure such as objects moving or appearing on the screen or perspective changing in multiple different realistic ways. 
We used the same architecture across all tasks, namely a 3-layer convolutional network followed by a 3-layer deconvolutional network, all with 64 feature maps at each layer and batch normalization. We did not use pooling and instead used strided convolutions, similar to the DCGAN architecture \citep{DCGAN}. The parametric function $\phi$ mapping the prediction error to latent variables was also a multilayer convolutional network followed by two fully-connected layers. For Atari Breakout we used 2 latent variables, for Seaquest, Flappy Bird and the Robot dataset we used 8, and for driving we used 32. To train our network we used the ADAM optimizer \citep{ADAM} with default parameters and learning rate 0.0005 for all tasks. 
The deterministic baseline model and the GAN had the same encoder-decoder architecture as the EEN, with twice as many feature maps. 

\begin{figure}[t]
\centering
\subfloat[a) Ground truth]{%
  \includegraphics[width=\textwidth]{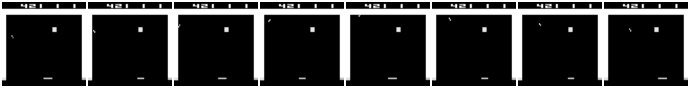}%
  }\par
\subfloat[b) Deterministic Baseline]{%
  \includegraphics[width=\textwidth]{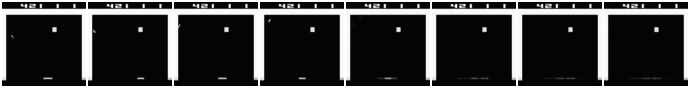}%
  }\par
\subfloat[c) Residual]{%
  \includegraphics[width=\textwidth]{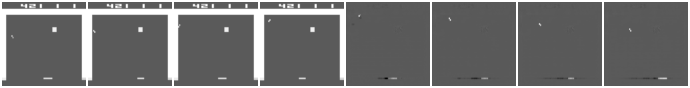}%
  }\par
\subfloat{%
  \includegraphics[width=\textwidth]{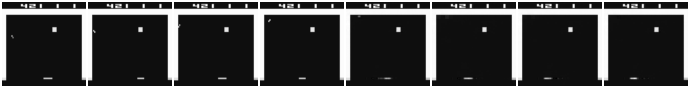}%
  }\par
\subfloat{%
  \includegraphics[width=\textwidth]{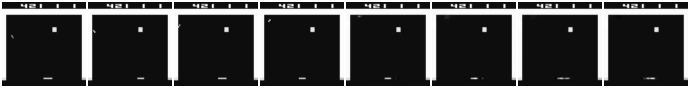}%
  }\par
\subfloat[d) Generations with different $z$]{%
  \includegraphics[width=\textwidth]{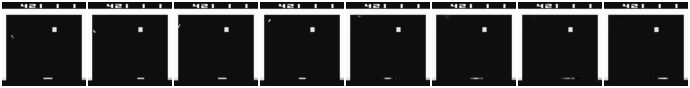}%
  }\par
\caption{Generations on Breakout. Left 4 frames are given, right 4 frames are generated. Note that the paddle changes location for the different generations. Best viewed with zoom.}
\label{breakout-img}
\end{figure}

\begin{figure}[t]
\centering
\subfloat[a) Ground Truth]{%
  \includegraphics[width=\textwidth]{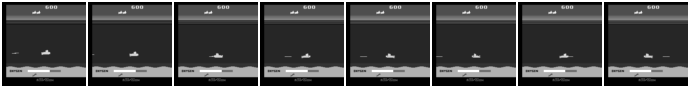}%
  }\par
\subfloat[a) Deterministic Baseline]{%
  \includegraphics[width=\textwidth]{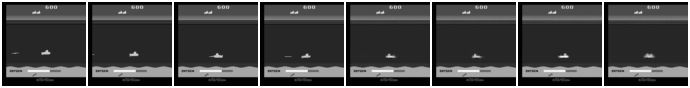}%
  }\par
\subfloat[c) Residual]{%
  \includegraphics[width=\textwidth]{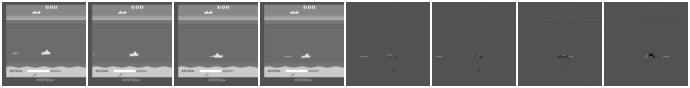}%
  }\par
\subfloat{%
  \includegraphics[width=\textwidth]{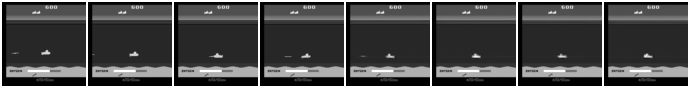}%
  }\par
\subfloat{%
  \includegraphics[width=\textwidth]{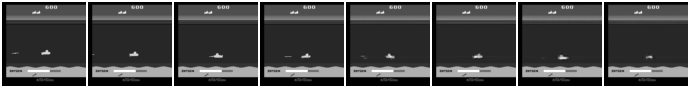}%
  }\par
\subfloat[b) Generations with different $z$]{%
  \includegraphics[width=\textwidth]{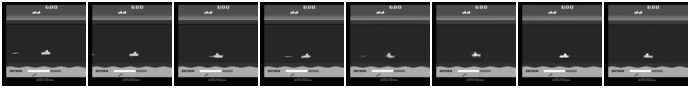}%
  }\par
\caption{Generations on Seaquest. Left 4 frames are given, right 4 frames are generated. Note that the submarine changes orientation for the different generations. Best viewed with zoom.}
\label{seaquest-img}
\end{figure}


\subsection{Datasets}

We now describe the video datasets we used. 

\textbf{Atari Games} We used a pretrained A2C agent \citep{A3C} \footnote{https://github.com/ikostrikov/pytorch-a2c-ppo-acktr} to generate episodes of gameplay for the Atari games Breakout and Seaquest \citep{Atari} using a standard video preprocessing pipeline, i.e. downsampling video frames to $84 \times 84$ pixels and converting to grayscale. We then trained our forward model using 4 consecutive frames as input to predict either the following 1 frame or 4 frames. 

\textbf{Flappy Bird} We used the OpenAI Gym environment Flappy Bird \footnote{https://gym.openai.com/envs/FlappyBird-v0/} and had a human player play approximately 50 episodes of gameplay. In this environment, the player controls a moving bird which must navigate between obstacles appearing at different heights. We trained the model to predict the next 4 frames using the previous 4 frames as input, all of which were rescaled to $128 \times 72$ pixel color images.

\textbf{Robot Manipulation} We used the dataset of \citep{Poke}, which consists of $240 \times 240$ pixel color images of objects on a table before and after manipulation by a robot. 
The robot pokes the object at a random location with random angle and duration causing it to move, hence the manipulation does not depend of the environment except for the location of the object. Our model was trained to take a single image as input and predict the following image. 

\textbf{Simulated Driving} We used the dataset from \citep{Jiakai2016}, which consists of color videos from the front of a car taken within the TORCS simulated driving environment. This car is driven by an agent whose policy is to follow the road and pass or avoid other cars while staying within the speed limit. Here we again trained the model to predict 4 frames using the 4 previous frames as input. Each image was rescaled to $160 \times 72$ pixels as in the original work.

\subsection{Results}

Our experiments were designed to test whether our method can generate multiple realistic predictions given the start of a video sequence. 
We first report qualitative results in the form of visualizations. In addition to the figures in this paper, we provide a link to videos which facilitate viewing \footnote{\url{www.mikaelhenaff.net/eenvideos.html}}.
An example of generated frames in Atari Breakout is shown in Figure \ref{breakout-img}. For the baseline model, the image of the paddle gets increasingly diffuse over time which reflects the model's uncertainty as to its future location while the static background remains well defined. The residual, which is the difference between the ground truth and the deterministic prediction, only depicts the movement of the ball and the paddle which the deterministic model is unable to predict. This is encoded into the latent variables $z$ through the learned function $\phi$ which takes the residual as input. By sampling different $z$ vectors from the training set, we obtain three different generations for the same conditioning frames. For these we see a well-defined paddle executing different movement sequences starting from its initial location. 

Figure \ref{seaquest-img} shows generations for Atari Seaquest. Again we see the baseline model captures most of the features on the screen except for the agent's movement, which appears in the residual. This is the information that will be encoded in the latent variables, and by sampling different latent variables we obtain the generations below where the submarine changes direction. 

\begin{figure}[t]
\centering
\subfloat[a) Deterministic Baseline]{%
  \includegraphics[width=0.5\textwidth]{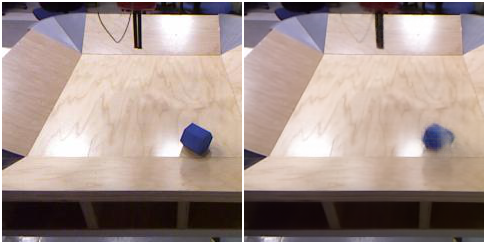}%
  }
\subfloat[b) Generation 1]{%
  \includegraphics[width=0.5\textwidth]{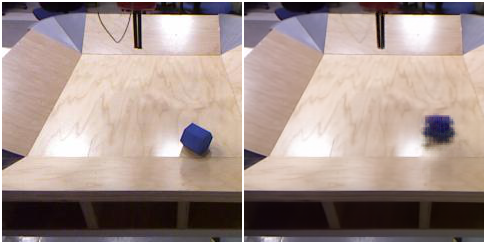}%
  }\par
\subfloat[c) Generation 2]{%
  \includegraphics[width=0.5\textwidth]{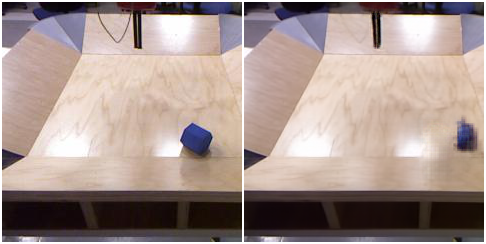}%
  }
\subfloat[d) Generation 3]{%
  \includegraphics[width=0.5\textwidth]{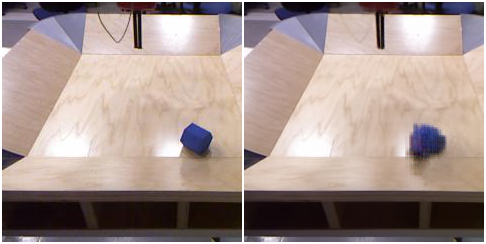}%
  }\par
\caption{Generations on Robot Task. Left frame is given, right frame is generated. The object is moved to different locations for the different generations.}
\vspace{-0.75cm}
\label{robot-img}
\end{figure}

We next evaluated our method on the Robot dataset. For this dataset the robot pokes the object with random direction and force which cannot be predicted from the current state. The prediction of the baseline model blurs the object but does not change its location or angle. In contrast, our model is able to produce a diverse set of predictions where the object is moved to different adjacent locations, as shown in Figure \ref{robot-img}. 

\begin{figure}
\centering
\subfloat[a) Deterministic Baseline]{%
  \includegraphics[width=0.45\textwidth]{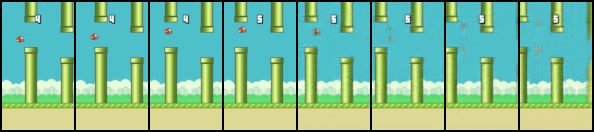}%
  }\hspace{0.01mm}
\subfloat[b) Generation 1]{%
  \includegraphics[width=0.45\textwidth]{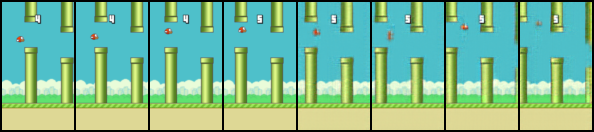}%
  }\par
\subfloat[c) Generation 2]{%
  \includegraphics[width=0.45\textwidth]{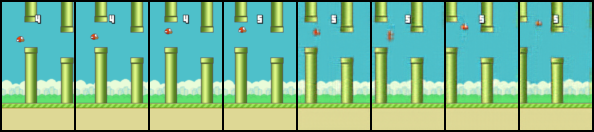}%
  }\hspace{0.01mm}
\subfloat[d) Generation 3]{%
  \includegraphics[width=0.45\textwidth]{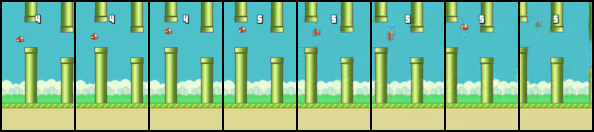}%
  }\par
\caption{Generated frames on Flappy Bird. First 4 are given, last 4 are generated. Note that the pipe in the last frame appears at different heights. Best viewed with zoom.}
\label{flappy-img1}
\centering
\subfloat[a) Deterministic Baseline]{%
  \includegraphics[width=0.45\textwidth]{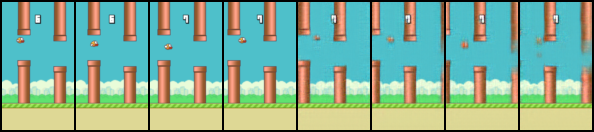}%
  }\hspace{0.01mm}
\subfloat[b) Generation 1]{%
  \includegraphics[width=0.45\textwidth]{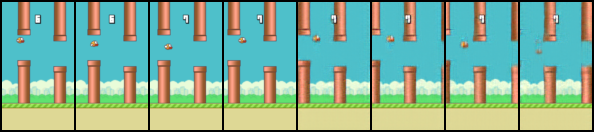}%
  }\par
\subfloat[c) Generation 2]{%
  \includegraphics[width=0.45\textwidth]{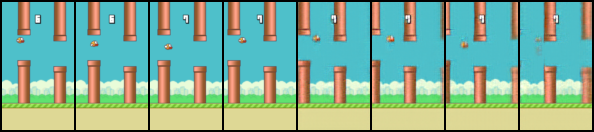}%
  }\hspace{0.01mm}
\subfloat[d) Generation 3]{%
  \includegraphics[width=0.45\textwidth]{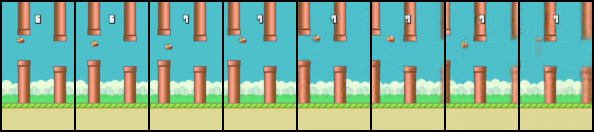}%
  }\par
\caption{Generated frames on Flappy Bird. First 4 are given, last 4 are generated. Note that the bird changes height in the last generated frame. Best viewed with zoom.}
\label{flappy-img2}
\end{figure}

Figures \ref{flappy-img1} and \ref{flappy-img2} show generated frames on Flappy Bird. Flappy Bird is a simple game which is deterministic except for two sources of stochasticity: the actions of the player and the height of new pipes appearing on the screen. In the first example, we see that by changing the latent variable we generate two sequences with pipes entering at different moments and heights and one sequence where no pipe appears. In the second example, changing the latent variable changes the height of the bird. The EEN is thus able to model both sources of uncertainty in the environment. Additional examples can be found at the provided video link. 

The last dataset we evaluated our method on was the TORCS driving simulator. Here we found that generating frames with different $z$ samples changed the location of stripes on the road, and also produced translations and dilations of the frame as would happen when turning the steering wheel or changing speed. These effects are best viewed though the video link. 

We next report quantitative results. Quantitatively evaluating multimodal predictions is not obvious, since the ground truth sample is drawn from one of several possible modes and the model may generate a sample from a different mode. In this case, simply comparing the generated sample to the ground truth sample may give high loss even if the generated sample is of high quality. We therefore report the best score across different generated samples: $\underset{k}{\mbox{ min }} \mathcal{L}(y, f(x, z_k))$. 
If the multimodal model is able to use its latent variables to generate predictions which cover several modes, generating more samples will improve the score since it increases the chance that a generated sample will be from the same mode as the test sample. 
If however the model ignores latent variables or does not capture the mode that the test sample is drawn from, generating more samples will not improve the loss. Note that if $\mathcal{L}$ is a valid metric in the mathematical sense (such as the $\ell_1$ or $\ell_2$ distance), this is a finite-sample approximation to the Earth Mover or Wasserstein-1 distance between the true and generated distributions on the metric space induced by $\mathcal{L}$.  

\begin{figure}
\centering
\subfloat{%
  \includegraphics[width=0.25\textwidth]{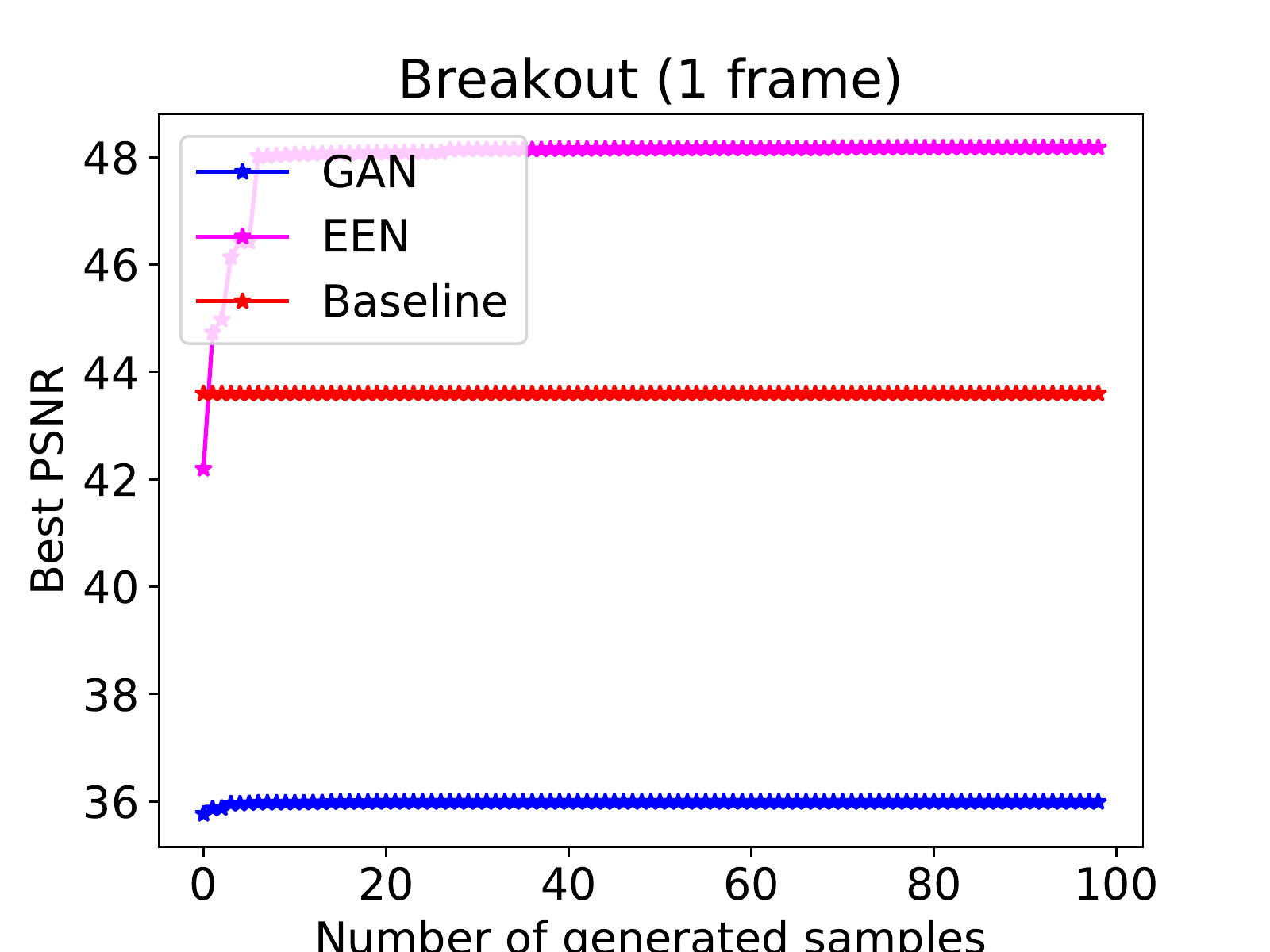}%
  }
\subfloat{%
  \includegraphics[width=0.25\textwidth]{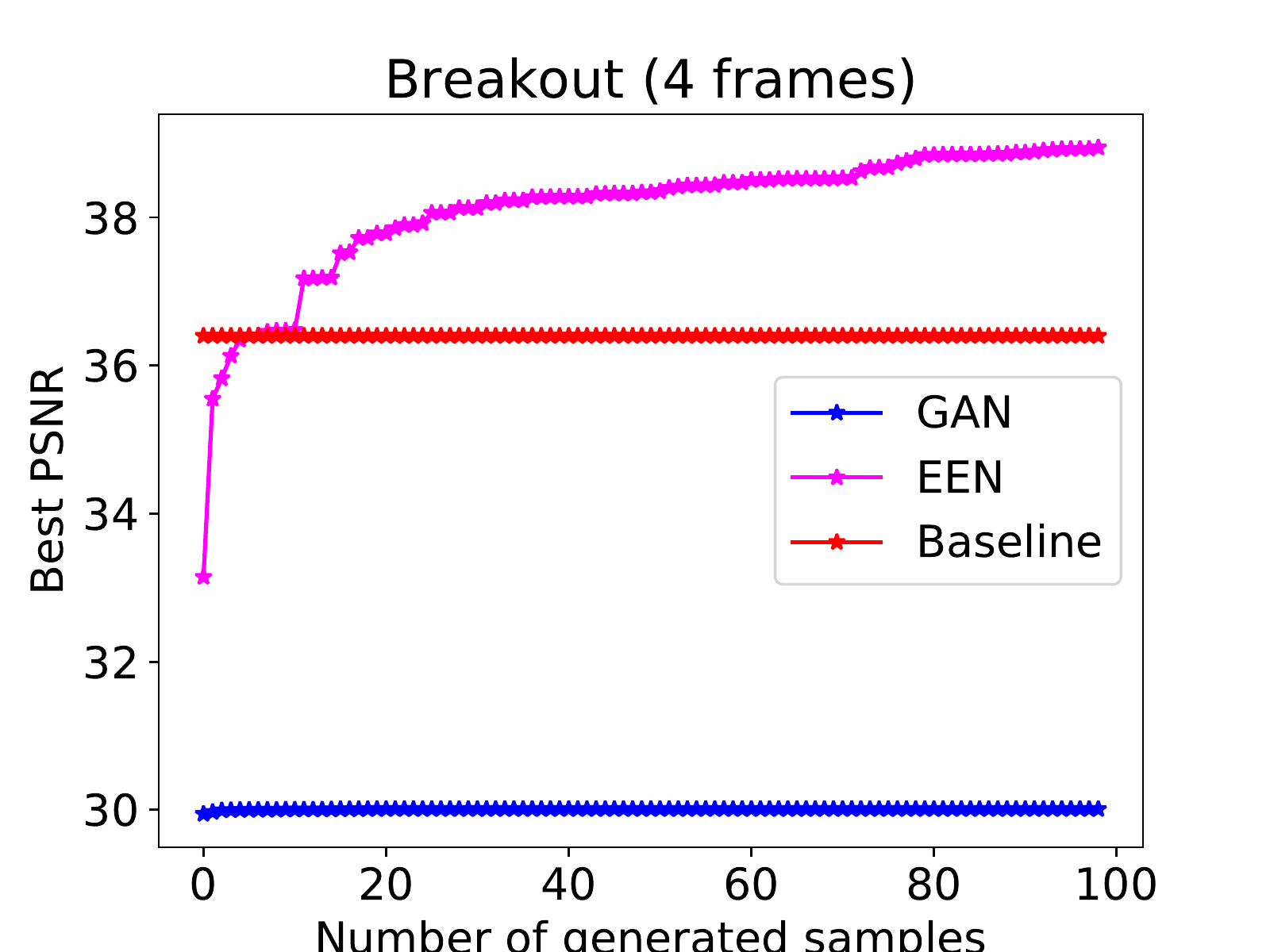}%
  }
\subfloat{%
  \includegraphics[width=0.25\textwidth]{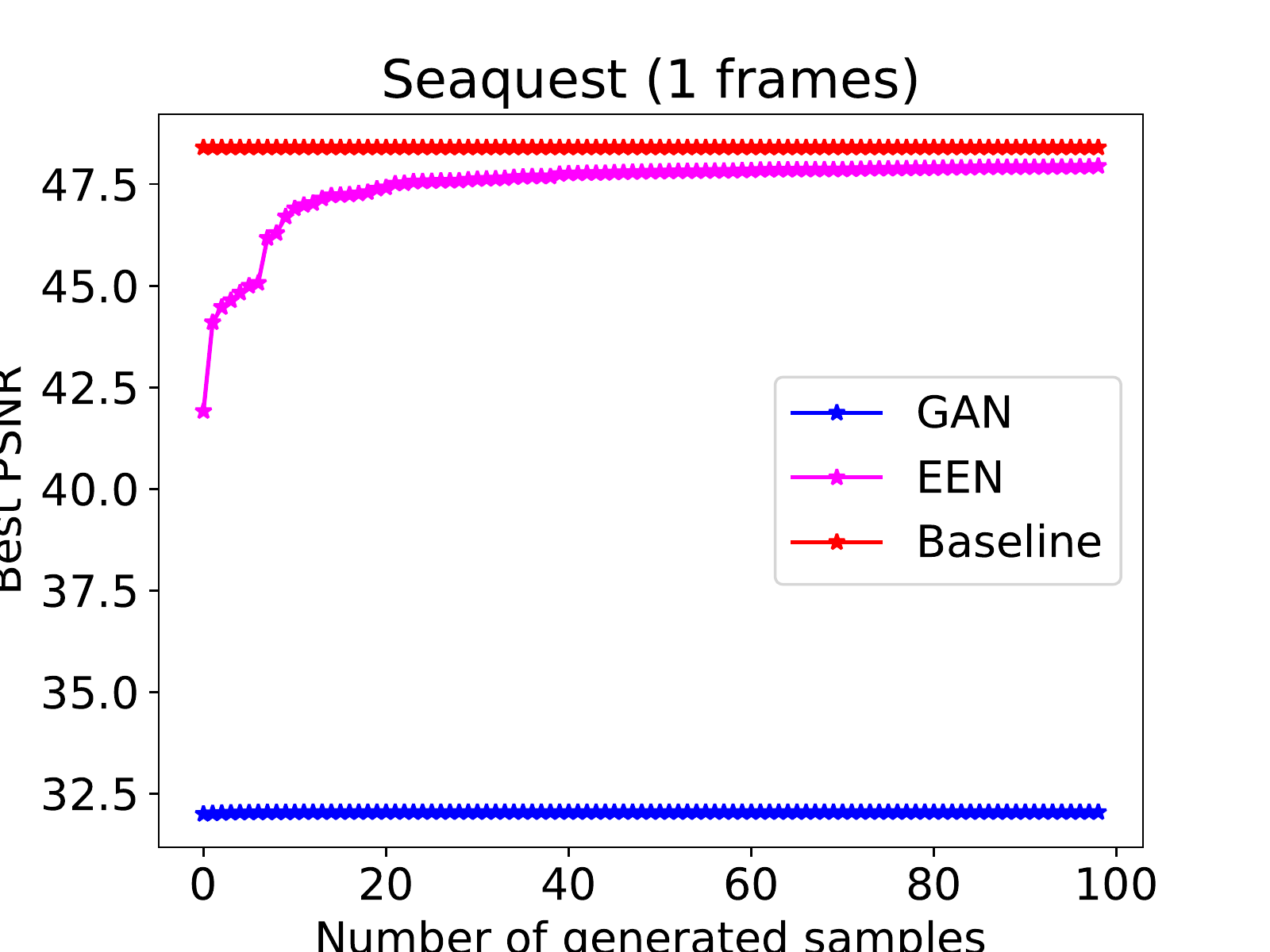}%
  }\par
\subfloat{%
  \includegraphics[width=0.25\textwidth]{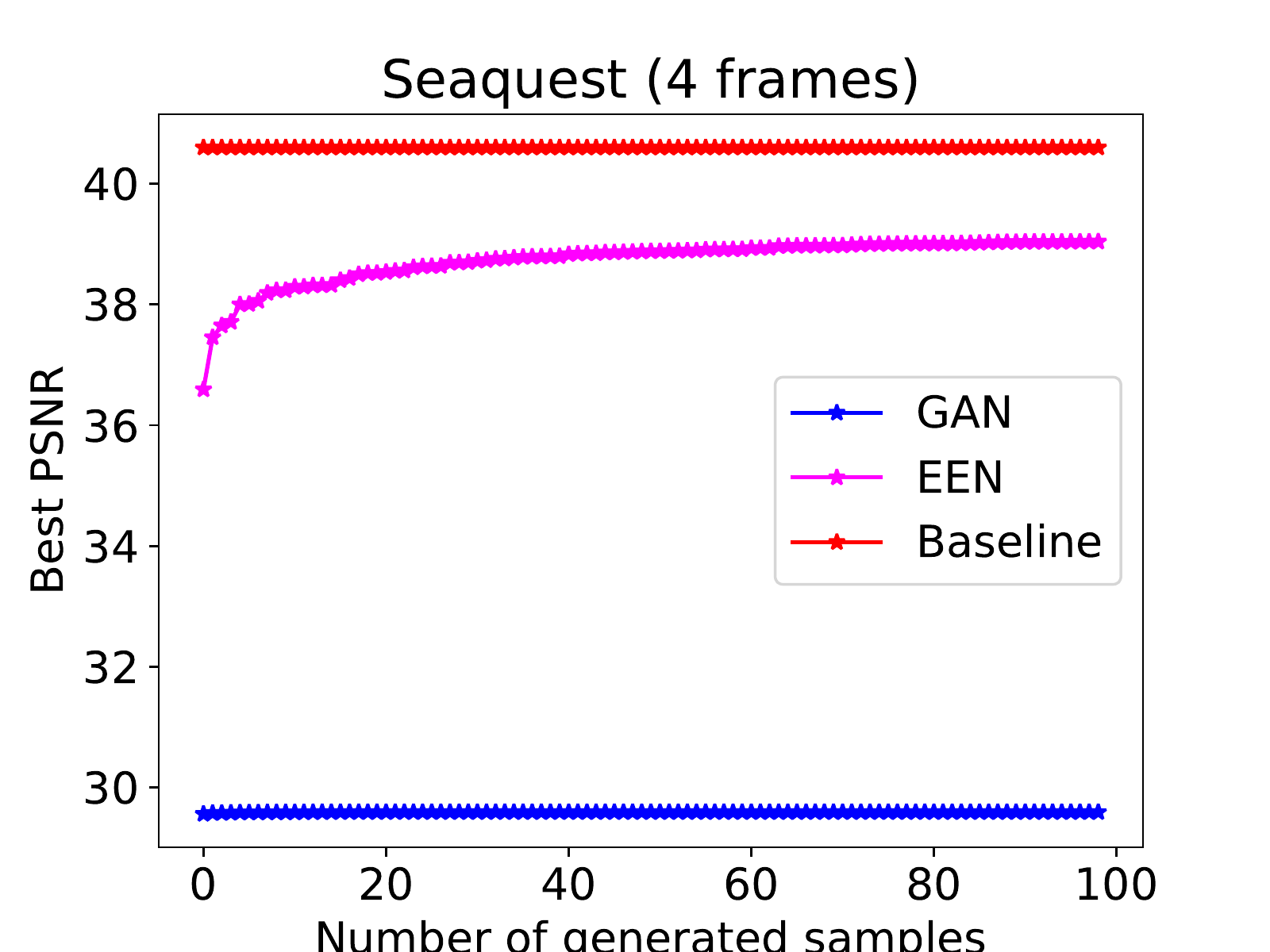}%
  }
\subfloat{%
  \includegraphics[width=0.25\textwidth]{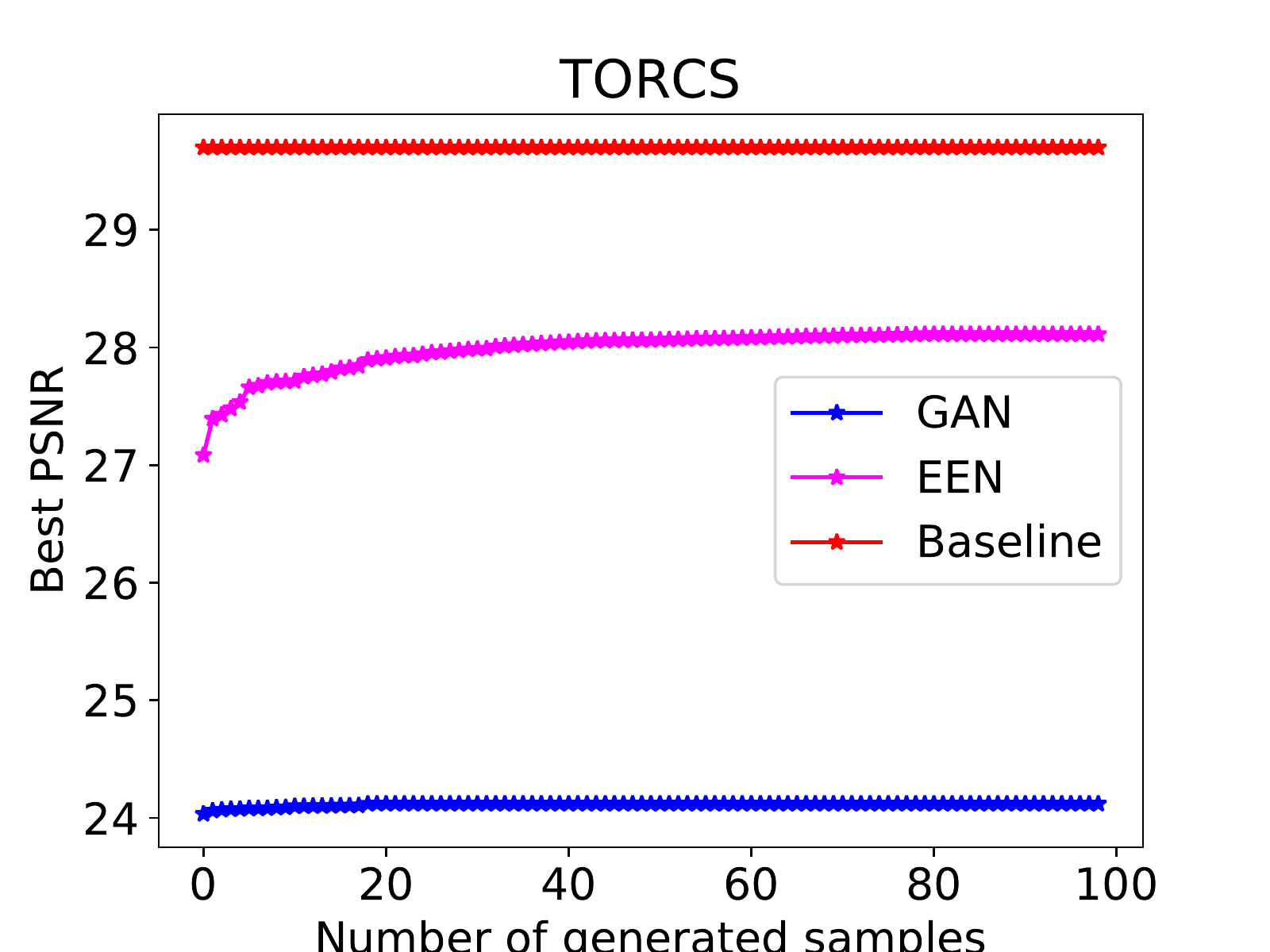}%
  }
\subfloat{%
  \includegraphics[width=0.25\textwidth]{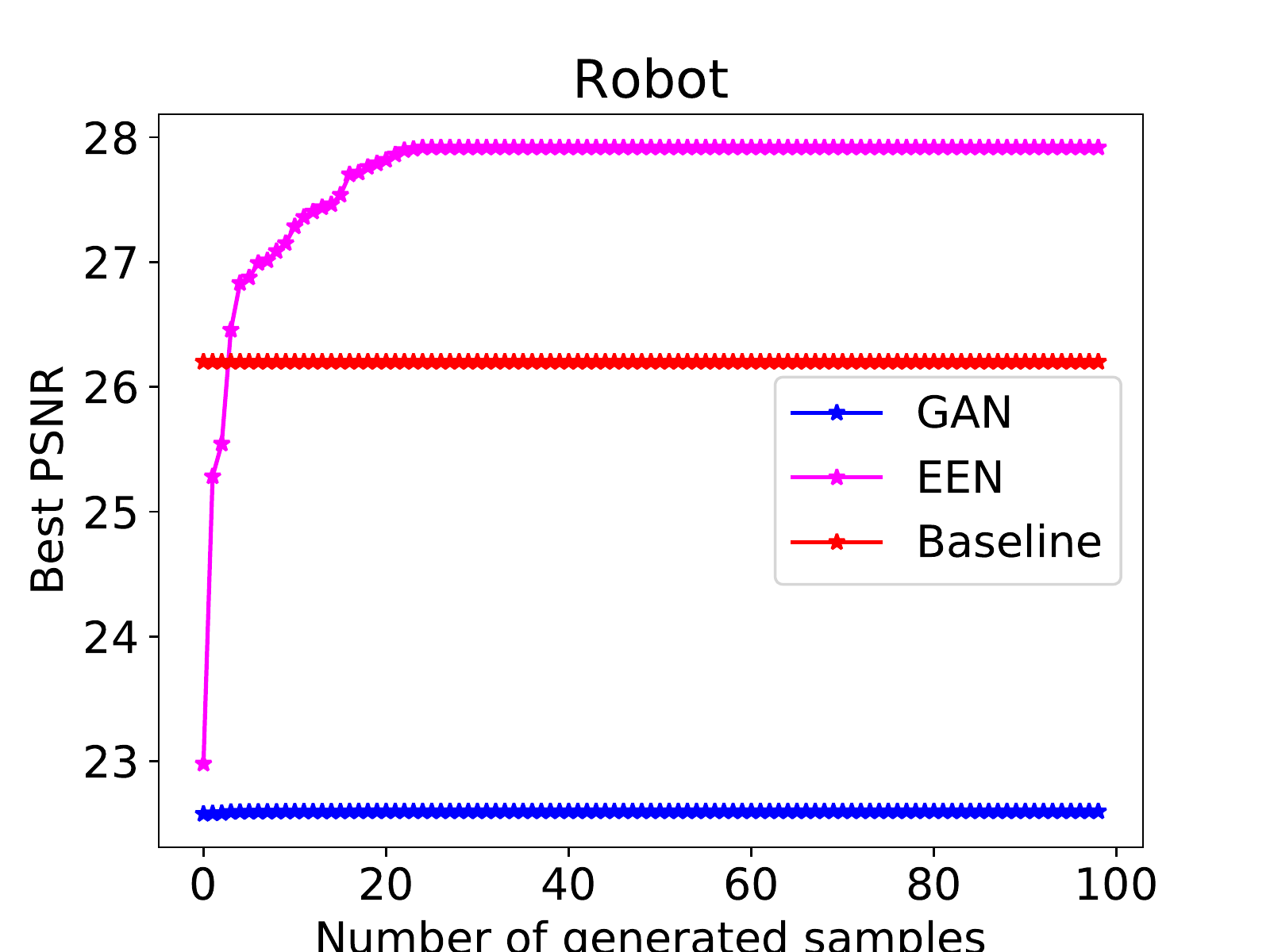}%
  }\par
\subfloat{%
  \includegraphics[width=0.25\textwidth]{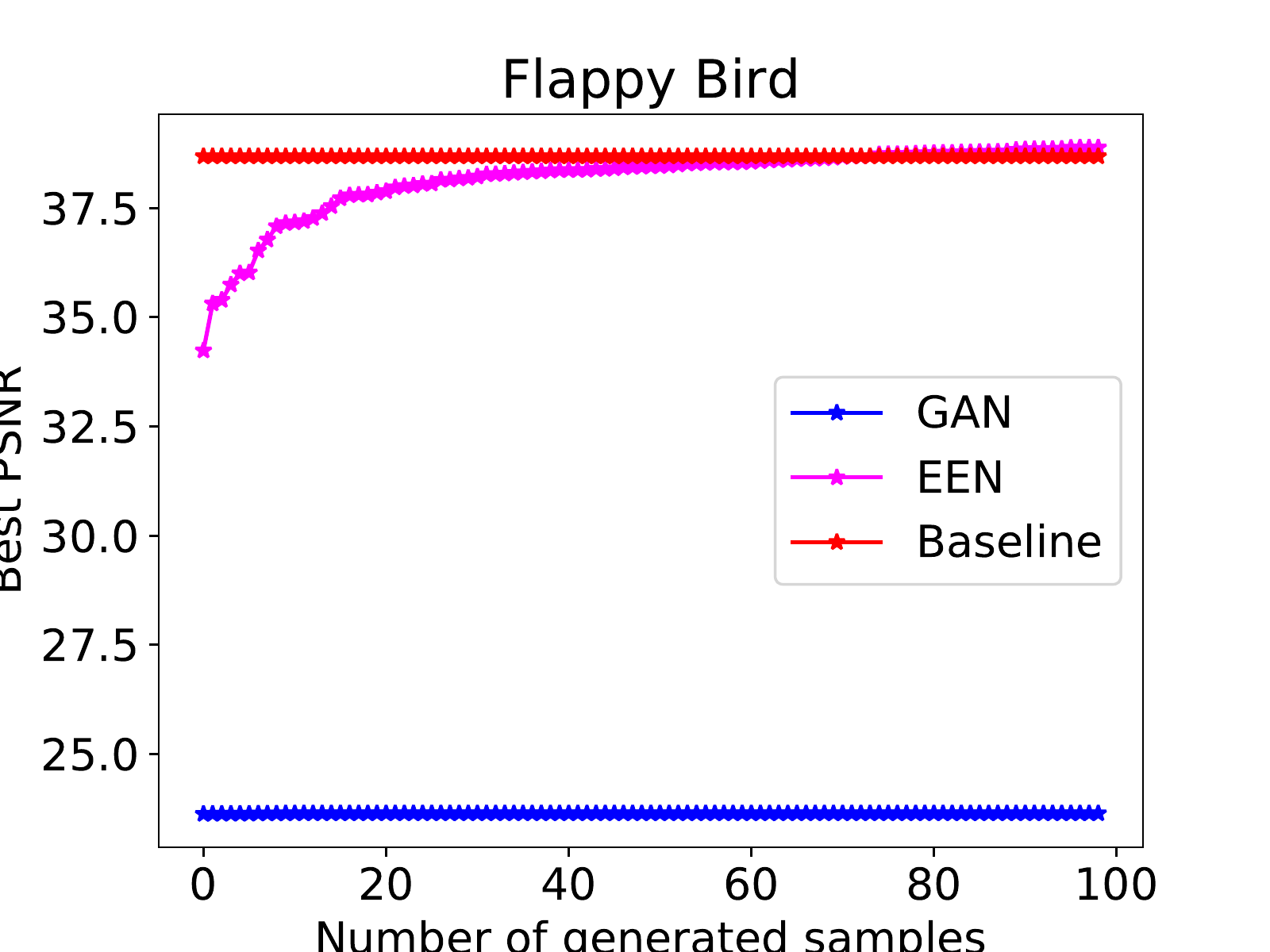}%
  }\par
\caption{Top PSNR for different models over varying numbers of different samples. The PSNR for the EEN increases with more samples, indicating it is able to generate predictions which span several modes, whereas the GAN does not. See text below.}
\label{psnr}
\end{figure}

Figure \ref{psnr} shows the best PSNR for different numbers of generated samples. 
For the Robot task, we report results for a model trained using the $\ell_2$ loss to make it consistent with the other models.
We see that our model's best performance increases as more samples are generated, indicating that its generations are diverse enough to cover at least some of the modes of the test set. Also note that the GAN's performance does not change as we increase the number of samples generated, which indicates that its latent variables have little effect on the generated samples. This is consistent with findings in other work \citep{Mathieu15, Isola2016}. 
We also note that the different models are not quite comparable to each other using PSNR since the baseline model is directly optimizing the $\ell_2$ loss on which it is based, the EEN is optimizing it conditioned on knowledge of a specific test sample, and the GAN is optimizing a different loss altogether. Our main goal is to illustrate that our model's performance improves by this approximate measure as it generates more samples, whereas the GAN does not due to mode collapse.

\section{Conclusion}

In this work, we have introduced a new framework for performing temporal prediction in the presence of uncertainty by disentangling predictable and non-predictable components of the future state. It is fast, simple to implement and easy to train without the need for an adverserial network or alternating minimization. 
We have provided one instantiation in the context of video prediction using convolutional networks, but it is in principle applicable to different data types and architectures. 
There are several directions for future work. Here, we have adopted a simple strategy of sampling uniformly from the $z$ distribution without considering their possible dependence on the state $x$, and there are likely better methods. In addition, one advantage of our model is that it can extract latent variables from unseen data very quickly, since it simply requires a forward pass through a network. If latent variables encode information about actions in a manner that is easy to disentangle, this could be used to extract actions from large unlabeled datasets and perform imitation learning. Another interesting application would be using this model for planning and having it unroll different possible futures.

\subsubsection*{Acknowledgments}

We would like to thank Jiakai Zhang and Kyunghyun Cho for sharing their dataset with us, and Martin Arjovsky, Arthur Szlam and Gabriel Synnaeve for helpful discussions.

\bibliography{iclr2017_conference}
\bibliographystyle{iclr2017_conference}

\end{document}